\documentclass[10pt,twocolumn,letterpaper]{article}

\usepackage{cvpr}
\usepackage{times}
\usepackage{epsfig}
\usepackage{graphicx}
\usepackage{amsmath}
\usepackage{amssymb}

\usepackage[pagebackref=true,breaklinks=true,letterpaper=true,colorlinks,bookmarks=false]{hyperref}

\usepackage{multirow}
\usepackage{amsthm}
\usepackage{bm}
\usepackage{times}
\usepackage{subfigure}
\usepackage{algorithm}
\usepackage{algorithmic}
\newtheorem{theorem}{Theorem}
\newtheorem{proposition}[theorem]{Proposition}

\usepackage{threeparttable}

\cvprfinalcopy 

\def\cvprPaperID{1298} 

\ifcvprfinal\fi
\begin{document}

\title{Fractal Dimension Invariant Filtering and Its CNN-based Implementation}

\author{Hongteng Xu$^{1,2}$,~Junchi Yan$^{3}$\thanks{corresponding author},~Nils Persson$^4$,~Weiyao Lin$^5$,~Hongyuan Zha$^{2}$\\
School of $^1$ECE, $^2$CSE, $^4$Chemical \& Biomolecular Engineering, Georgia Tech\\
$^3$IBM Research -- China, $^5$Department of EE, Shanghai Jiao Tong University\\
{\tt\small $\{$hxu42,~npersson3$\}$@gatech.edu,~yanjc@cn.ibm.com,~wylin@sjtu.edu.cn,~zha@cc.gatech.edu}
}

\maketitle
\begin{abstract}
Fractal analysis has been widely used in computer vision, especially in texture image processing and texture analysis. 
The key concept of fractal-based image model is the fractal dimension, which is invariant to bi-Lipschitz transformation of image, and thus capable of representing intrinsic structural information of image robustly. 
However, the invariance of fractal dimension generally does not hold after filtering, which limits the application of fractal-based image model. 
In this paper, we propose a novel fractal dimension invariant filtering (FDIF) method, extending the invariance of fractal dimension to filtering operations. 
Utilizing the notion of local self-similarity, we first develop a local fractal model for images.
By adding a nonlinear post-processing step behind anisotropic filter banks, we demonstrate that the proposed filtering method is capable of preserving the local invariance of the fractal dimension of image.
Meanwhile, we show that the FDIF method can be re-instantiated approximately via a CNN-based architecture, where the convolution layer extracts anisotropic structure of image and the nonlinear layer enhances the structure via preserving local fractal dimension of image. 
The proposed filtering method provides us with a novel geometric interpretation of CNN-based image model. 
Focusing on a challenging image processing task --- detecting complicated curves from the texture-like images, the proposed method obtains superior results to the state-of-art approaches. 

\end{abstract}

\section{Introduction}
Many complex natural scenes can be modeled as fractals~\cite{mandelbrot1983fractal,pentland1984fractal}.
In the field of computer vision, fractal analysis has been proven to be a useful tool for modeling textures, and many research fruits have been proposed.
Taking textures as fractals, the work in~\cite{varma2007locally} learns local fractal dimensions and lengths as features for classifying textures.
Similarly, the work in~\cite{xu2009viewpoint} learns the spectrum of fractal dimension as textures' features via the box-counting method~\cite{falconer2004fractal}.
It is easy to find that all of these methods treat the fractal dimension as a key concept of fractal-based image model because the fractal dimension is invariant to bi-Lipschitz transformation. This property means that the fractal dimension is robust to geometrical deformation (e.g., ridge and non-ridge transformation) of image. Hence, the fractal dimension reflects intrinsic structural information of image, which can be treated as a representative feature of image.
\begin{figure}[!tb]
\begin{center}
\subfigure[]{
\includegraphics[width=0.11\columnwidth]{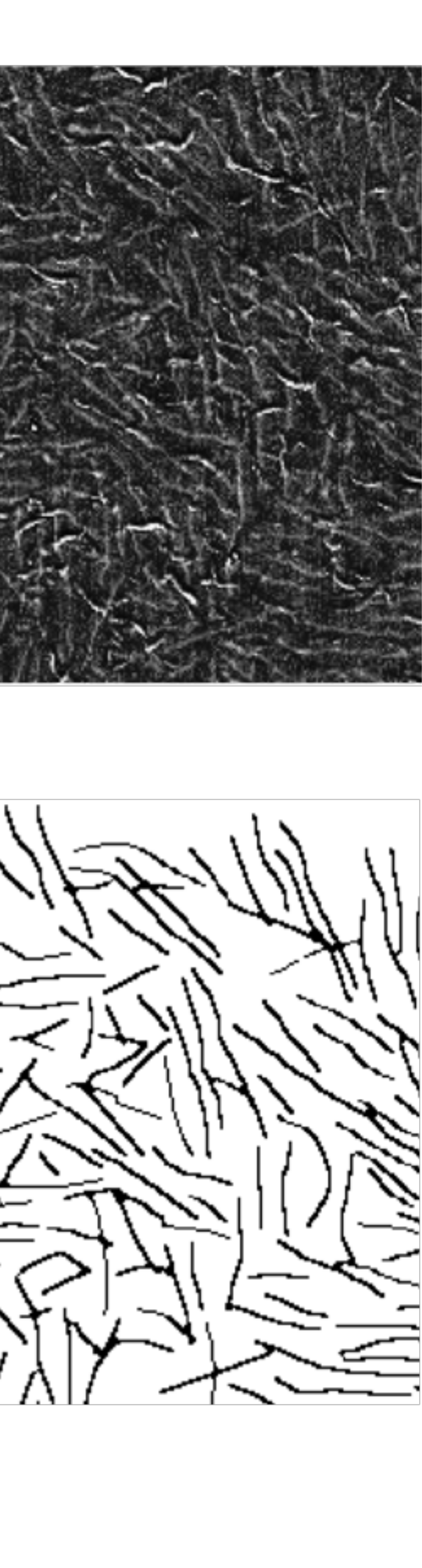}\label{Fig1a}
}
\subfigure[Iterative FDIF]{
\includegraphics[height=3.5cm]{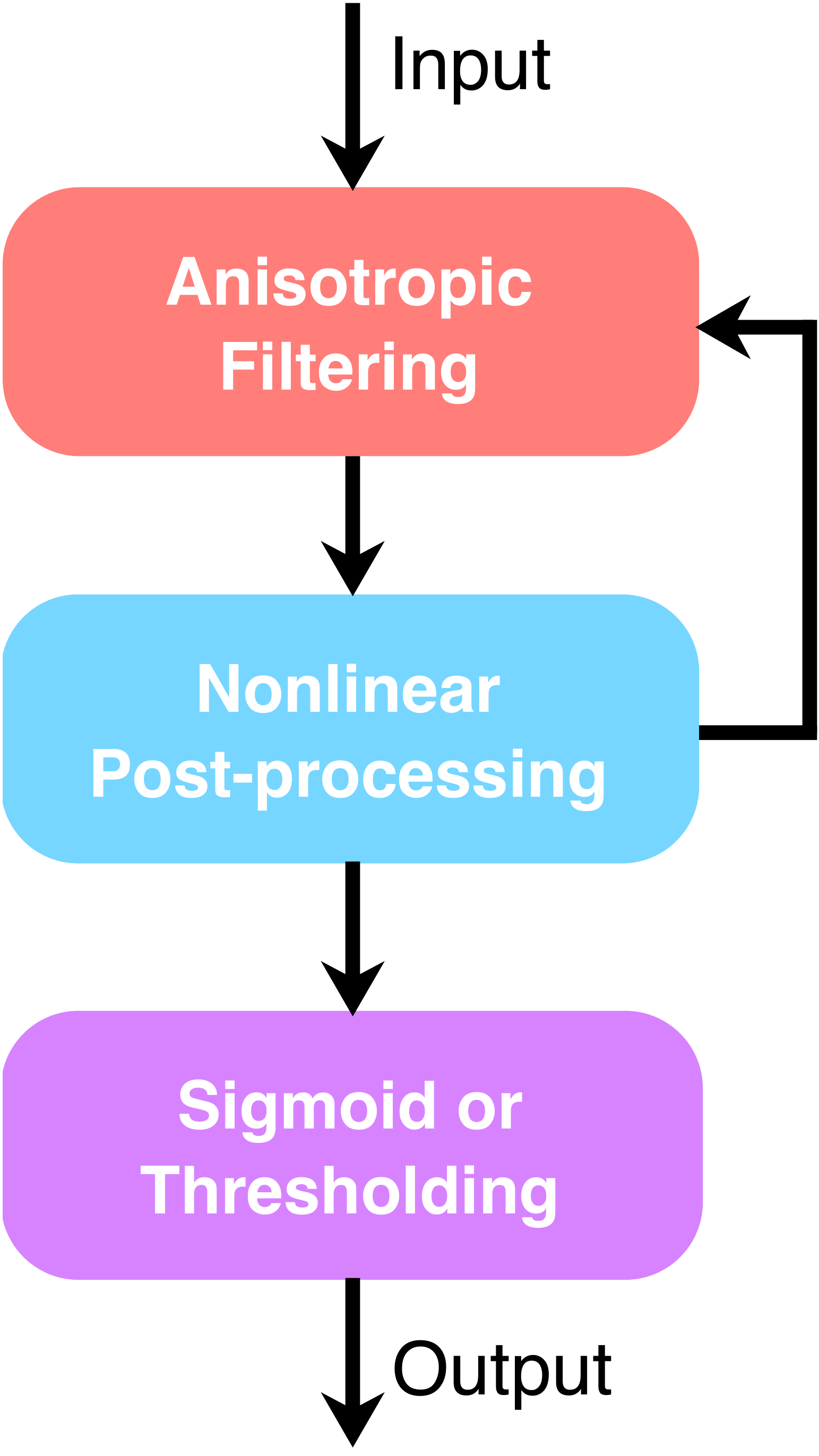}\label{Fig1b}
}
\subfigure[CNN-based Implementation]{
\includegraphics[height=3.5cm]{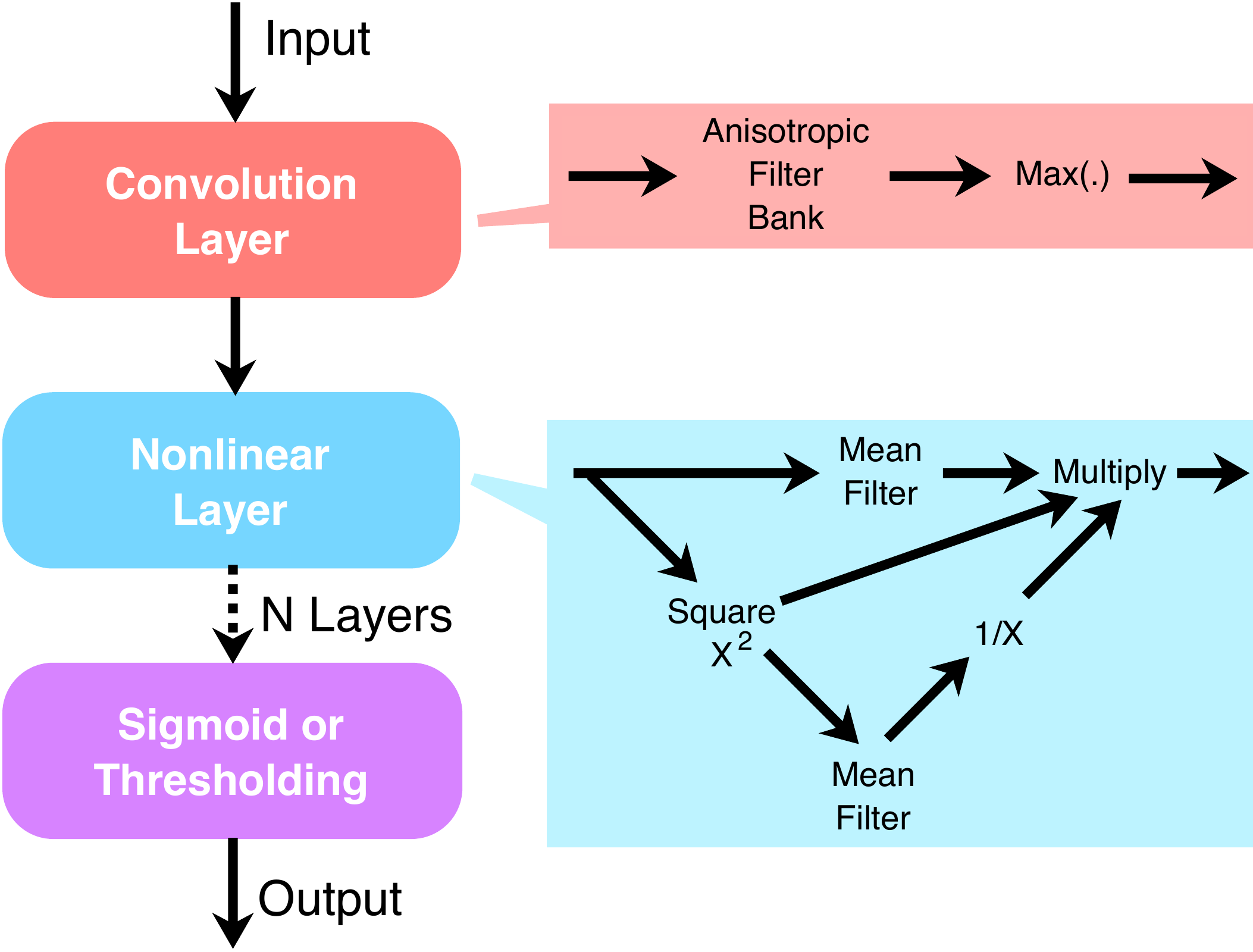}\label{Fig1c}
}
\subfigure[Illustration of the FDIF-based Curve Detector]{
\includegraphics[width=1\columnwidth]{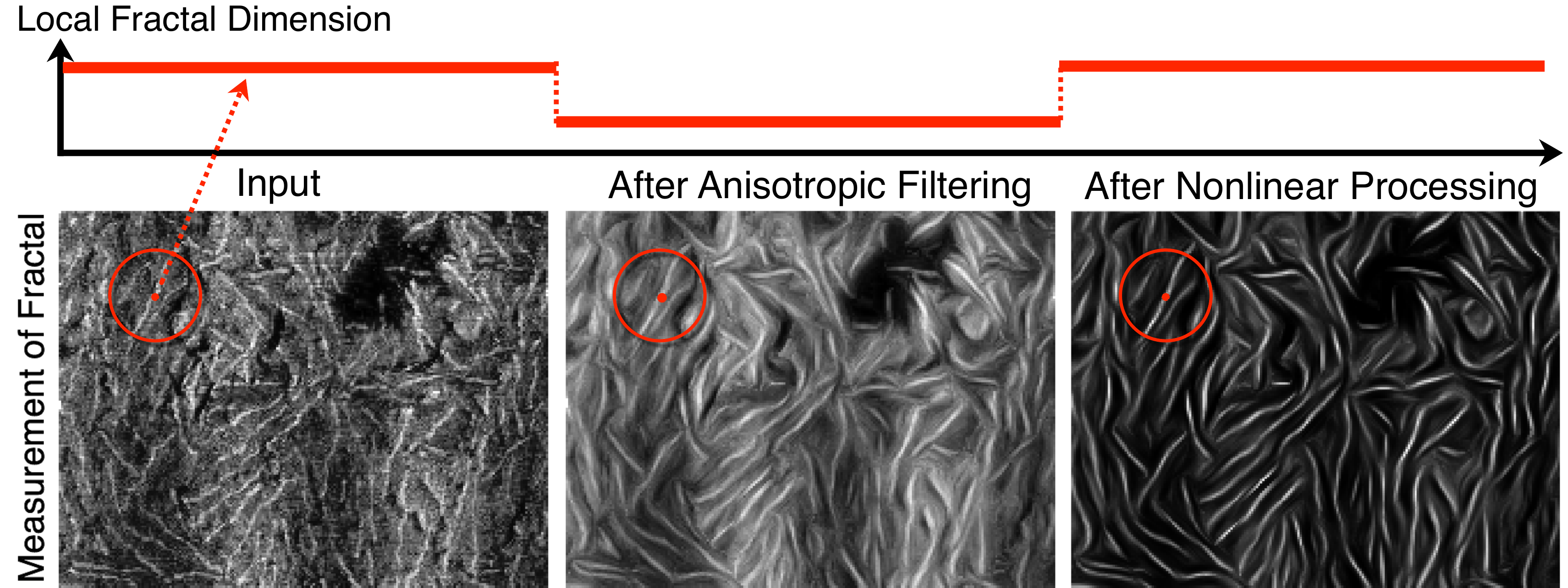}\label{Fig1d}
}\vspace{-10pt}
\end{center}
   \caption{Given real-world noisy images (i.e., material images) having complicated curves in (a), we apply the proposed iterative FDIF method in (b) to detect curves. The FDIF can be efficiently and approximately re-instantiated via a CNN in (c). The illustration of the FDIF-based curve detector is shown in (d).}
\label{fig1}
\end{figure}

Unfortunately, the fractal dimension of image cannot be preserved after filtering, which might lead to the loss of structural information.
A typical example is the interpolation of digital image, where the result can be viewed as a low-pass filtering of ground truth.
The low-pass filtering suppresses the high-resolution details of image, and thus, leads to the loss of structural information.
The work in~\cite{xu2013single} shows that the fractal dimension of interpolated image is smaller than that of real high-dimensional image.
However, the recent development of deep convolutional neural networks (CNNs) shows that the stacked nonlinear filtering model is very suitable to learn features of images, which has a capability of extracting structural and semantic information of image robustly.
Many CNN-based methods have been proposed to deal with various tasks e.g. image classification~\cite{he2015deep}, texture analysis~\cite{bruna2013invariant}, and contour detection~\cite{yang2016object}.
{In other words, for extracting representative feature of image, the filtering operation are instrumental in CNNs while detrimental to fractal-based methods.}
Given these two seemingly contradictory phenomena, the following two problems arise:
1) Can we propose a filtering method preserving the invariance of fractal dimension?
2) Is there any connection between fractal-based image models and CNNs, especially for unsupervised feature learning?

In this paper, we give positive answers to these two problems.
\emph{We propose a fractal dimension invariant filtering (FDIF) method and use a CNN-based architecture to re-instantiate it.
This work provides us with a geometrical interpretation of CNN based on local fractal analysis of image.
The proposed work obtains encouraging curve detection results for texture-like images, which is superior to other competitors.}
As Fig.~\ref{Fig1b} shows, we give a local fractal model of image and propose a curve detector under an iterative FDIF framework.
In each iteration, we take patches of image as local fractals, and compute their fractal dimensions accordingly.
An anisotropic filter is designed for each patch of image according to the analysis of gradient field, and the filtering result is further enhanced via preserving fractal dimension across various measurements.
Inspired by the iterative filtering strategy in~\cite{milanfar2013tour}, we apply the steps above repeatedly to obtain the features of curves, and detect curves via unsupervised (i.e., thresholding) or supervised (i.e., logistic regression) methods.
In particular, we demonstrate that such a pipeline can be implemented via a CNN-based architecture, as shown in Fig.~\ref{Fig1c}.
This CNN is interpretable from a geometrical viewpoint --- the convolution layer  corresponds to an anisotropic filter bank while the nonlinear layer approximately preserves local fractal dimensions.
Applying backpropagation algorithm for supervised case and predefined parameters (filters) for unsupervised case, we achieve encouraging curve detection results.

{As Fig.~\ref{Fig1d} shows, the principle of our FDIF-based curve detector is preserving local fractal dimensions via adjusting the measurement of fractal (i.e., the image itself).
Generally, the measurement obtained via anisotropic filtering is smoothed.
To preserve local fractal dimensions, we apply the nonlinear processing and get a new measurement, where the sharpness of curve is enhanced while the sharpness of the rest regions is suppressed.
As a result, the FDIF method provides us with a better representation of curves.}

We test our method on a collected atomic-force microscopy (AFM) image set, detecting complicated curves of materials from AFM images.
Experimental results show that our method is promising in most situations, especially in the noisy and texture-like cases, which obtains superior results to existing curve detectors.
Overall, the contributions of our work are mainly in three aspects:
First, to the best of our knowledge, our work is the first attempt to propose a fractal dimension invariant filtering method and connect it with CNNs.
It is also perhaps the first time to interpret CNNs from a (fractal) geometry perspective.
Second, our method connects traditional handcrafted filter-based curve detector with a CNN architecture.
It establishes a bridge on the gap between filter-based curve detectors and learning-based especially CNN-based ones.
This connection also allows us to instantiate a new predefined CNN that can work in an unsupervised setting, different from most of its peers known for their ravenous appetite for labeled data.
Third, we demonstrate a meaningful interdisciplinary application of our curve detector in computational material science.
A material informatics image dataset is collected and will be released with this paper for future public research.

\section{Related Work}
\textbf{Fractal Analysis:} Fractal-based image model has been widely used to solve many problems of computer vision, including, texture analysis~\cite{quan2014lacunarity}, bio-medical image processing~\cite{wang2016dynamic}, and image quality assessment~\cite{xu2015fractal}.
The local fractal analysis method in~\cite{varma2007locally} and the spectrum of fractal dimension in~\cite{zhang2016block,xu2009viewpoint} take advantage of the bi-Lipschitz invariance property of fractal dimension for texture classification, whose features are very robust to the deformation and scale changing of textures.
Because the local self-similarity of image is often ubiquitous both within and across scales~\cite{glasner2009super,freedman2011image}, natural images can also be modeled as fractals locally~\cite{mandelbrot1983fractal,pentland1984fractal}.
Recently, the fractal model of natural image is applied to image super-resolution~\cite{xu2013single,licheng2013self}, where the local fractal analysis is used to enhance image gradient adaptively.
In~\cite{wang2016dynamic}, a fracal-based dissimilarity measurement is proposed to analyze MRI images.
However, because the invariance of fractal dimension does not hold after filtering, it is difficult to merge fractal analysis into other image processing methods.

\textbf{Convolution Neural Networks:} CNNs have been widely used to extract visual features from images, which have many successful applications.
In these years, this useful tool has been introduced into many low-and middle-level vision problems, e.g., image reconstruction~\cite{xie2012image,burger2012image}, super-resolution~\cite{dong2014learning}, dynamic texture synthesis~\cite{yan2014modeling}, and contour detection~\cite{xie2015holistically,yang2016object}.
Currently, the physical meanings of different CNN modules are not fully comprehended.
For example, the nonlinear layer of CNN, i.e., the rectifier linear unit (ReLU), and its output are often mysterious.
For comprehending CNNs in depth, many attempts have been made.
Many existing feature extraction methods have been proven to be equivalent to deep CNNs, like deformable part models in~\cite{girshick2015deformable} and random forests in~\cite{patel2015probabilistic}.
A pre-trained deep learning model called scattering convolution network (SCN) is proposed in~\cite{mallat2012group,bruna2013invariant,Oyallon_2015_CVPR}.
This model consists of hierarchical wavelet transformations and translation-invariant operators, which explains deep learning from the viewpoint of signal processing.
However, none of these methods discuss the geometrical explanation of CNNs from the viewpoint of fractal analysis.

\textbf{Curve Detection:}
Curve detection is a potential application of fractal-based image processing method regarding many practical tasks, such as power line detection~\cite{ma2011algorithm}, geological measurement~\cite{park2003lane}, and rigid body detection~\cite{puatruaucean2012parameterless} etc. More recently, the curve detection technique is introduced into more interdisciplinary fields, e.g., materials, biology, and nanotechnology~\cite{yang2000three,takacs2013remarkable,jordens2013non}.
To our surprise, although in the following section we show that fractal-based image model is very suitable for the problem of curve detection, very few existing methods apply fractal analysis to solve the problem.
Taking advantage of the directionality of curve, early curve detectors are based on diverse transformations, including the Hough transformation~\cite{duda1972use}, the curvelets~\cite{starck2002curvelet}, the wave atoms~\cite{xu2012automatic}.
Besides the direction, the multiscale property of curve is considered via applying multiscale Fourier transformation~\cite{calway1990curve}, Frangi filtering~\cite{frangi1998multiscale}, and the scale-space distance transformation~\cite{sironi2014multiscale}. 
Focusing on curve and line segment detection, the parameterless fitting model proposed in~\cite{puatruaucean2012parameterless} achieves the state-of-the-art. 
These methods principally construct an isotropic filter bank and detect the local strong response to certain directions. 
Beyond these manually-designed methods, the learning-based approaches become popular as a huge amount of labeled images become available~\cite{arbelaez2011contour,zhang2012contour}.
Focusing on edge detection, which is a problem related to curve detection, the structured forest-based detector~\cite{dollar2015fast} and the CNN-based detector~\cite{shen2015deepcontour,bertasius2015deepedge,xie2015holistically,shen2015shadow} are proposed.
These methods learn their parameters on a large dataset, and thus, have powerful generalization ability to deal with challenging cases.
However, most of the existing methods aim to detect sparse curves from relatively smooth background.
Few of them can detect complicated curves from texture-like images.

\section{Fractal Dimension Invariant Filtering}
\label{sec:blind}

In this section, we introduce our fractal-based image model and show the derivation of local fractal dimension.
According to the model, we propose an iterative fractal dimension invariant filtering method, which preserves local fractal dimensions of patches across various measurements in the phase of feature extraction.

\subsection{Fractal-based Image Model}
\begin{figure}[!t]
\begin{center}
\includegraphics[width=1\linewidth]{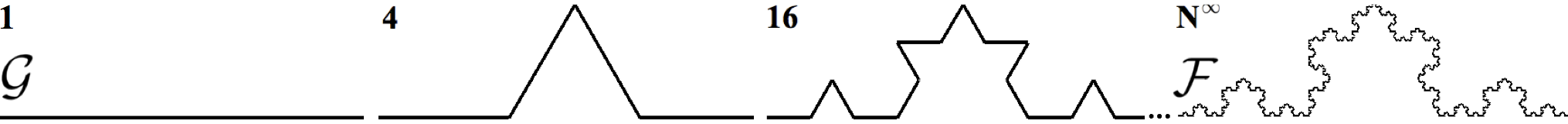}\vspace{-10pt}
\end{center}
   \caption{Transforming each line segment into $N=4$ analogues with scaling factor $s=\frac{1}{3}$ repeatedly, we obtain a fractal so-called the Von Koch curve with $D=\frac{\log N}{-\log s}=1.262$.
   }
\label{fig2}
\end{figure}

As shown in Fig.~\ref{fig2}, a typical fractal is generated via transforming a geometry $\mathcal{G}$ to $N$ analogues with scaling factor $s$ and then applying the transformation infinitely on each analogue.
The union of the analogues is a fractal, denoted as $\mathcal{F}$.
The fractal $\mathcal{F}$ is a ``Mathematical monster'' that is unmeasurable in the measure space of $\mathcal{G}$.
Therefore, the analysis of fractal is mainly based on the Hausdorff measure~\cite{mandelbrot1983fractal}, which gives rise to the concept of fractal dimension.
The fractal dimension is involved by a power law of measurements across multiple scales, i.e., the quantities $N \varpropto \frac{1}{s^D}$.
Here $D$ is called \emph{fractal dimension}, which is larger than the topological dimension of $\mathcal{F}$.

In our work, an image is represented via a function of pixels, denoted as $f(\bm{X})$.
Here $\bm{X}\subset \mathbb{R}^2$ is the union of the coordinates of pixels.
Each coordinate of pixel is denoted as $\bm{x}\in\bm{X}$.
We propose a fractal-based image model, representing $\bm{X}$ as a union of local fractals, and image $f(\bm{X})$ as ($\bm{X}$, $\mu$), where $\mu$ is a measurement supported on the fractal set $\bm{X}$.
According to the power law of measurements mentioned above, for each pixel $\bm{x}$ we have $\mu(B_{r}(\bm{x}))\varpropto (2r)^{D(\bm{x})}$, where $B_{r}(\bm{x})$ is a ball centering at $\bm{x}$ with radius $r$ and $D(\bm{x})$ is the local fractal dimension at $\bm{x}$ under the measurement $\mu$.
Here, we use the intensity of pixel $f(\bm{X})$ as the measurement $\mu$ directly, so the local fractal dimension at $\bm{x}$ is
\begin{eqnarray}\label{fractaldim}
\begin{aligned}
D(\bm{x})=\lim_{r\rightarrow 0}\frac{\log\mu(B_{r}(\bm{x}))}{\log 2r},
\end{aligned}
\end{eqnarray}
where $\mu(B_{r}(\bm{x}))=\int_{\bm{y}\in B_{r}(\bm{x})} G_{r}*f(\bm{y}) d\bm{y}$, $G_{r}=\frac{\exp(-x^2/r^2)}{\sqrt{2\pi}r}$ is a Gaussian kernel defined as~\cite{xu2009viewpoint,xu2013single}, and ``$*$'' indicates the valid convolution. 

In practice, we estimate the local fractal dimension in~(\ref{fractaldim}) numerically by linear regression.
Specifically, we calculate sample pairs $\{\log r, \log\mu(B_{r}(\bm{x}))\}_{r=\{1,2,...\}}$ by multiscale Gaussian filtering, and learn a linear model $\log\mu(B_{r}(\bm{x}))=D(\bm{x})\log{2r} +L(\bm{x})$ for all $\bm{x}\in\bm{X}$ according to~(\ref{fractaldim}).
Here $\exp(L(\bm{x}))$ is the value of measurement $\mu$ in the unit ball ($2r=1$), which is interpreted as the $D$-dimensional fractal length in~\cite{varma2007locally}.
Algorithm~\ref{alg1} gives the scheme of fractal dimension estimation.
\begin{algorithm}[h]
   \caption{Fractal Dimension Estimation}
   \label{alg1}
\begin{algorithmic}[1]
   \STATE \textbf{Input:} $f(\bm{X})$, the number of scales $R$.
   \STATE \textbf{Output:} Fractal dimension $D(\bm{X})$.
   \STATE For $r\in\{1,...,R\}$, perform a convolution of $f(\bm{X})$ with $G_r$ to get $\{\mu(B_{r}(\bm{x}))\}_{\bm{x}\in\bm{X}}$.
   \STATE $\min_{D,L}\sum_{r}|\log\mu(B_{r}(\bm{x}))-D\log{2r} -L|^2$, $\bm{x}\in\bm{X}$.
   \STATE $D(\bm{X})=\{D(\bm{x})\}_{\bm{x}\in\bm{X}}$.
\end{algorithmic}
\end{algorithm}

Local Fractal dimension contains important structural information of image, e.g., smooth patches with fractal dimensions close to $2$, the patch containing curves with fractal dimensions close to $1$, and textures with fractal dimensions between $1$ and $2$~\cite{falconer2004fractal,xu2013single}.
For detecting structures, e.g., curves in images robustly, fractal dimension shall be preserved. One fundamental property of fractal dimension is its invariance to bi-Lipschitz transform shown in Theorem~\ref{the1}:
\begin{theorem}\label{the1}
\textbf{Bi-Lipschitz Invariance.}
For a fractal $\mathcal{F}$ with fractal dimension $D$, its bi-Lipschitz transformation $g(\mathcal{F})$ is still a fractal, whose fractal dimension $D_g=D$.
\end{theorem}
Recall~(\ref{fractaldim}), we can find that the fractal dimension is not unique, which depends on the choice of measurement $\mu$.
The theorem holds because the bi-Lipschitz transformation (i.e., the geometric transformation and non-rigid deformation of image) does not change the measurement of fractal, which is revealed via the proof in the appendix.

However, after filtering or convolution, the invariance of fractal dimension does not hold any more.
For example, if we change the convolution kernel $G_r$ in~(\ref{fractaldim}), the measurement $\mu$ of fractal $\bm{X}$ and the associated fractal dimension will be changed accordingly.
Therefore, we cannot find a filter ensuring the fractal dimension of filtering result to be exactly same with that of original image.

To pursuit the fractal dimension preservation philosophy in face of the reality that filtering will inevitably change fractal dimension, we aim to suppress the expected change between original fractal dimension and filtered one.
Denote the proposed filter as $F$, the measurement and the fractal dimension of filtering result as $\mu_{F}$ and $D_{F}$, respectively.
We assume that the filter $F$ is a random variable yielding to a probabilistic distribution.
According to~(\ref{fractaldim}), we have
\begin{eqnarray}\label{condition}
\begin{aligned}
&D(\bm{x})-\mathbb{E}(D_F(\bm{x}))
=\lim_{r\rightarrow 0}\frac{1}{\log 2r}\log\frac{\mu(B_{r}(\bm{x}))}{\mathbb{E}(\mu_D(B_{r}(\bm{x})))}\\
&=\lim_{r\rightarrow 0}\frac{1}{\log 2r}\log\frac{\int_{\bm{y}\in B_{r}(\bm{x})} G_{r}*f(\bm{y}) d\bm{y}}{\int_{\bm{y}\in B_{r}(\bm{x})} G_{r}*\mathbb{E}(F)*f(\bm{y}) d\bm{y}},
\end{aligned}
\end{eqnarray}
where $\mathbb{E}(\cdot)$ computes the expectation of random variable.
Obviously, to minimize the expected change between $D(\bm{x})$ and $D_F(\bm{x})$, the expectation of the filter should be as close to impulse function  $\delta(\bm{x})$ as possible.

\subsection{Iterative FDIF Framework}\label{sec:filter}
Motivated by the analysis above, we propose the following iterative FDIF method as detailed in Fig.~\ref{Fig1b}.

\textbf{Anisotropic Filtering:} To suppress fractal dimension change, the expectation of the filter shall be as close as to impulse function.
Anisotropic filters have been one natural choice for this purpose. Take directional filtering~\cite{peyre2010texture} as an example:
for each pixel $\bm{x}$, compute the smoothed gradient in its neighborhood $B(\bm{x})$ as $\bm{G}=[\mbox{vec}(\nabla_{h}(G*f(B(\bm{x}))), \mbox{vec}(\nabla_{v}(G*f(B(\bm{x})))]\in \mathbb{R}^{|B|\times 2}$.
Here $G$ is a Gaussian filter, $|B|$ is the cardinality of the neighborhood, $\nabla_h$ ($\nabla_v$) is partial differential operator along horizontal (vertical) direction, and $\mbox{vec}(\cdot)$ denotes vectorization.
The eigenvector corresponding to the largest eigenvalue of $\bm{G}^{\top}\bm{G}$, denoted as $\bm{u}=[u_h,u_v]^{\top}\in \mathbb{R}^{2}$, indicates the direction information of $\bm{x}$.
Such a direction field of image induces a series of directional filters in the polar coordinate system, denoted as $F_{\theta}$, whose element $F_{\theta}(r,\phi)$ satisfies
\begin{eqnarray}\label{filter}
\begin{aligned}
F_{\theta}(r,\phi)=
\begin{cases}
\frac{1}{|B|},& \mbox{$\phi\in\{\theta, \theta+\pi\}$, $r\in[0,\sqrt{|B|}]$},\\
0, &\mbox{otherwise}.
\end{cases}
\end{aligned}
\end{eqnarray}
Obviously, the filtering result $f_F(\bm{x})=F_{\theta}*f(B(\bm{x}))$ at $\bm{x}$ has the strongest response for $\theta=\arctan\left(u_h/u_v\right)$.
The directional filters satisfy the following proposition:
\begin{proposition}\label{thm2}
If the distribution of pixel's direction is uniform, then the expectation value of the filters in~(\ref{filter}) is an impulse function $\delta(\bm{x})$, where $\delta(\bm{0})=\frac{1}{|B|}$.
\end{proposition}
The proof is given in the appendix. Fig.~\ref{fig3} visualizes several typical directional filters and their mean in the right most, which further verifies the proposition.
Recall~(\ref{condition}), we can find that as long as the distribution of directions is uniform in the direction field of image, the proposition indicates that the proposed filters $\{F_\theta\}$ tend to preserve the expected value of fractal dimension after filtering.
\begin{figure}[!t]
\begin{center}
\includegraphics[width=1\linewidth]{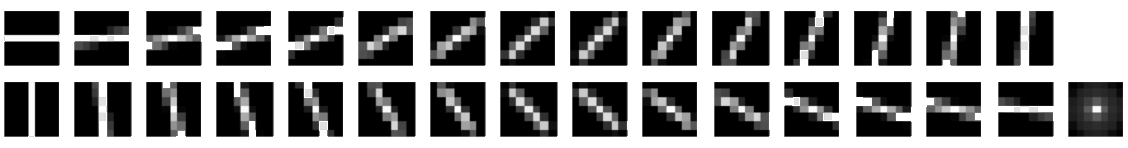}\vspace{-10pt}
\end{center}
   \caption{ The illustration of $F_{\theta}$'s with $\theta=\{0, \frac{\pi}{30}, ...,\frac{29\pi}{30}\}$. 
   The average of the filters (the last one) is close to an impulse function.
}
\label{fig3}
\end{figure}

\textbf{Nonlinear Post-processing:}
Anisotropic filtering prevents the expected fractal dimension from changing globally.
Furthermore, we propose a transformation $T$ to preserve local fractal dimensions of the filtering result $f_F(\bm{X})$.
In particular, although the local fractal dimension $D_F(\bm{x})$ with the measurement $\mu_F(B_r(\bm{x}))$ is not equal to the original $D(\bm{x})$ with $\mu(B_r(\bm{x}))$, we can apply a transformation $T$ to $\mu_F(B_r(\bm{x}))$, such that the fractal dimension with the new measurement $T(\mu_F(B_r(\bm{x})))$, denoted as $D_{T\circ F}(\bm{x})$, is equal to $D(\bm{x})$.
According to the definition of fractal dimension in~(\ref{fractaldim}) and the relationship $\log\mu(B_{r}(\bm{x}))=D\log{2r} +L$ given by Algorithm~\ref{alg1}, it is easy find that the proposed transformation should be $T=(\cdot)^{\alpha(\bm{x})}$, where $\alpha(\bm{x})=\frac{D(\bm{x})}{D_F(\bm{x})}$.
In this situation, we have
\begin{eqnarray*}\label{enhance}
\begin{aligned}
\log T(\mu_F(B_{r}(\bm{x})))
&=\frac{D(\bm{x})}{D_F(\bm{x})}\left(D_F(\bm{x})\log2r+L_F(\bm{x})\right)\\
&=D(\bm{x})\log2r+\frac{L_F(\bm{x})}{D_F(\bm{x})}.
\end{aligned}
\end{eqnarray*}
In other words, the local fractal dimension $D_{T\circ F}(\bm{x})=D(\bm{x})$.
Then we apply the transformation directly to the filtering result $f_F(\bm{X})$ such that the local fractal dimension is preserved under the new measurement.
At each $\bm{x}$, we have
\begin{eqnarray}\label{trans}
\begin{aligned}
f_{T\circ F}(\bm{x})=\frac{\|f_F(B(\bm{x}))\|}{\|f_F^{\alpha}(B(\bm{x}))\|}f_F^{\alpha}(\bm{x}),\quad\alpha=\frac{D(\bm{x})}{D_F(\bm{x})}.
\end{aligned}
\end{eqnarray}
Here the term $\frac{\|f_F(B(\bm{x}))\|}{\|f_F^{\alpha}(B(\bm{x}))\|}$ preserves the energy of filtering result, which merely changes fractal length.

\textbf{Iterative Framework:}
Combining the anisotropic filtering with the post-processing, we obtain the proposed FDIF method.
As Fig.~\ref{Fig1b} shows, FDIF can be applied iteratively, in order to extract structures hidden in images.

Take curve detection as an example.
Fig.~\ref{fig4} illustrates the enlarged output of an AFM image in each iteration and compare the iterative filtering process with traditional path operator~\cite{merveille2014tubular}. 
We can find that the pixels corresponding to curves are more and more discriminative.
When the labels of curves are available, we learn the curve detector as a binary classifier with the help of logistic regression.
Sampling the final filtering result into patches with overlaps, we learn the parameters of the sigmoid function.
On the contrary, if the labels are unavailable, we simply apply a thresholding method~\cite{otsu1975threshold} to convert the filtering result to a binary image. 
On the contrary, the traditional morphological filtering method, e.g., the path operator~\cite{talbot2007efficient,merveille2014tubular}, also aims at detecting curves and tubes, but it is sensitive to the noisy in the image. 
These two detection methods are shown in the last layer in Fig.~\ref{Fig1b}.
The iterative FDIF-based curve detector is physically-interpretable.
The fractal dimension of patch reflects its sharpness: the patch of curve has higher sharpness than the patch of smooth region, whose fractal dimension tends to $1$.
The filters we used achieve an anisotropic smoothing process of image, so that the measurement of fractal dimension is smoothed as well.
{Essentially, preserving fractal dimensions under a smoother measurement, like~(\ref{trans}) does, actually enhances the sharpness of curves and suppresses the sharpness of the rest regions, which provides us with a better representation of curves.}

\begin{figure}[!t]
\begin{center}
\subfigure[Original]{
\includegraphics[width=0.22\linewidth]{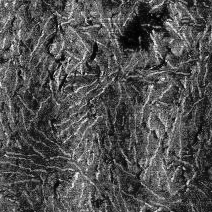}\label{Fig4a}
}
\subfigure[Path operator]{
\includegraphics[width=0.22\linewidth]{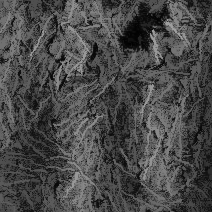}\label{Fig4b}
}
\subfigure[\#1 Iteration]{
\includegraphics[width=0.22\linewidth]{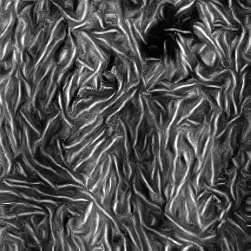}\label{Fig4c}
}
\subfigure[\#3 Iteration]{
\includegraphics[width=0.22\linewidth]{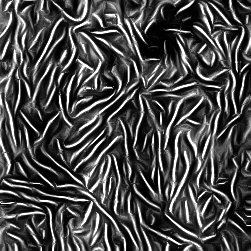}\label{Fig4d}
}\vspace{-10pt}
\end{center}
   \caption{Comparison between the iterative adaptive filtering process and traditional path operator~\cite{merveille2014tubular}.}
\label{fig4}
\end{figure}

\section{FraCNN: Implementing FDIF via CNN}
In this section, we will show that FDIF can be re-instantiated via a CNN, as described in Fig.~\ref{Fig1c}. 
In particular, the convolution layer can be explained as an anisotropic filter bank and the nonlinear layer performs the post-processing function approximately.
\subsection{The Architecture of The CNN}
\textbf{Convolution Layer:} The anisotropic filtering can be approximately implemented via a filter bank.
At each pixel $\bm{x}$, the process can be rewritten as
\begin{eqnarray}\label{conv}
\begin{aligned}
f_F(\bm{x})=F_{\theta}*f(B(\bm{x}))
=\max_{\theta\in\Theta}\{F_{\Theta}*f(B(\bm{x}))\},
\end{aligned}
\end{eqnarray}
where $F_{\Theta}=\{F_{\theta_1},...,F_{\theta_N}\}$ is the bank of $N$ anisotropic filters.
$\max_{\theta\in\Theta}\{F_{\Theta}*f(B(\bm{x}))\}$ only preserves the filtering result having the maximum response.

\textbf{Nonlinear Layer:}
The proposed post-processing can also be approximated via the following nonlinear layer:
\begin{eqnarray}
\begin{aligned}\label{nonlinear}
f_{T\circ F}(\bm{x})&\approx
\frac{\|f_F(B(\bm{x}))\|\max\{f_F(\bm{x}), 0\}^{\alpha}}{\|\max\{f_F(B(\bm{x}), 0\}^{\alpha}\|}\\
&=\frac{(M*f_F(B(\bm{x})))\max\{f_F(\bm{x}), 0\}^{\alpha}}{M*\max\{f_F(B(\bm{x})), 0\}^{\alpha}}.
\end{aligned}
\end{eqnarray}
Here the normalization term is implemented via a convolution, where $M$ is a mean filter, which sums the intensities in the neighborhood $B(\bm{x})$ for each $\bm{x}$.
Different from neuroscience, we explain the rectified linear unit (ReLU, $\max\{\cdot, 0\}$) based on fractal analysis. The ReLU ensures the filtering result to be a valid measurement (as the measurement used in the box-counting method~\cite{falconer2004fractal,xu2009viewpoint}): A valid measurement $\mu$ defined on the set $\bm{X}$ satisfies nonnegativity $\mu(X)\geq 0$, countable additivity $\mu(\cup_{k=1}^{\infty}X_k)=\sum_{k=1}^{\infty}\mu(X_k)$, and null empty set $\mu(\emptyset)=0$ simultaneously, where $X, X_k\subset \bm{X}$.
The null empty set is satisfied by our filtering result naturally while the ReLU operator guarantees the nonnegativity and countable additivity.

Note that the parameter of transformation operation $T(\cdot)=(\cdot)^{\alpha(\bm{x})}$ can be fixed approximately as a constant $\alpha$.
This approximation is reasonable for the problem of curve detection.
On one hand, we model the coordinates of image $\bm{X}$ as a set of fractals, whose fractal dimension $D(\bm{x})$ must be in the interval $[2,2+\epsilon_1]$, where $2$ is the topology dimension of 2D geometry, and $0\leq \epsilon_1<1$ because the fractal dimension of a fractal generated from a 2D geometry via 2D transformation cannot reach to $3$.
On the other hand, after filtering the curves are also modeled as a set of fractals with fractal dimension $D_{F}(\bm{x})$ in the interval $[1, 1+\epsilon_2)$, where $1$ is the topology dimension of curve (1D geometry) and $0\leq \epsilon_2<1$.
Based on the fractal-based model, we have $\alpha(\bm{x})=\frac{D(\bm{x})}{D_F(\bm{x})}\in (\frac{2}{1+\epsilon_2},2+\epsilon_1)$.
When $\epsilon_1$ and $\epsilon_2$ are small, we can estimate $\frac{D}{D_F}\approx 2$ for all $\bm{x}$'s.

\subsection{FraCNN-based Curve Detection}
The iterative FDIF framework can be achieved via stacking the layers above.
As a result, the architecture of the proposed CNN is shown in Figs.~\ref{Fig1c}.
For convenience, we call it \textbf{FraCNN}. Similar to the iterative FDIF framework, we can also add a sigmoid layer to the end of the CNN and train the model via traditional backpropagation algorithm, or apply a thresholding layer for the final output.
In contrast to many CNN models with a disadvantage of their ravenous appetite for labeled training, we believe the adaptability for unlabeled data of our method is perhaps due to the fact that we instantiate our tailored CNN from the fractal-based geometry perspective.
Focusing on the task of curve detection, we propose a detection algorithm shown in Algorithm~\ref{alg2}.
\begin{algorithm}[tb]
   \caption{FraCNN-based Curve Detector}
   \label{alg2}
\begin{algorithmic}[1]
   \STATE \textbf{Input:} Image $f(\bm{X})$, filter bank $F_{\Theta}$, layer number $N$.
   \STATE \textbf{Output:} Binary map $b(\bm{X})$ corresponding to curves.
   \STATE For $n=1,...,N$, obtain $f_{T\circ F}(\bm{X})$ from $f(\bm{X})$ via (\ref{conv},\ref{nonlinear}), and set $f(\bm{X})=f_{T\circ F}(\bm{X})$.
   \STATE \textbf{Unsupervised:} $b(\bm{X})=\mbox{binary}(f(\bm{X}))$.
   \STATE \textbf{Supervised:} $b(\bm{X})=\mbox{sigmoid}(\bm{\beta}^{\top}\bm{P})$. $\bm{\beta}$ is learned parameters, $\bm{P}$ are patch matrix of $f(\bm{X})$.
\end{algorithmic}
\end{algorithm}

We present further comparisons and analysis as follows.

\textbf{FraCNN v.s. FIDF:} The proposed CNN model can be viewed as a fast implementation of FIDF.
Firstly, the adaptive anisotropic filtering is approximately achieved by an anisotropic filter bank.
The direction of filter $\theta$ is no longer computed from the eigenvector of the local gradient matrix, but sampled uniformly from the interval $[0,\pi]$ (as Fig.~\ref{fig3} shows).
Although such an approximation reduces the accuracy of the description of direction, it avoids to do eigen-decomposition for each pixel, and thus, accelerates the filtering process notably.
Secondly, the ratio between the fractal dimension and the original one is replaced by a fixed value, such that we do not need to apply Algorithm~\ref{alg1} to estimate fractal dimension.
As a result, the computational complexity of original FIDF is $\mathcal{O}(|\bm{X}||B|^3+|\bm{X}|R^3)$, where the first term corresponds to adaptive filtering and the second term corresponds to local fractal dimension estimation (and $R$ is the number of scales in Algorithm~\ref{alg1}), while the complexity of proposed CNN is at most $\mathcal{O}(|\bm{X}||B|L)$, where $L$ is the number of filters in the filter bank.

\textbf{FraCNN v.s. Scattering Convolution Network:} To our CNN model, the most related work might be the scattering convolution network (SCN) in~\cite{bruna2013invariant,Oyallon_2015_CVPR}.
Both of our fractal-based CNN and the SCN can apply predefined filters and are suitable for unsupervised learning when labels are not available.
However, there are several important differences between our model and SCNs.
First, SCNs aim at extracting discriminative feature for image recognition and classification, while our Fractal-based CNN model focuses on low-and middle-level vision problems, i.e., curve detection.
Second, the nonlinear layer of SCN applies multiple nonlinear operators to enhance the invariance of feature to geometric transformation.
For example, the absolute operator $|\cdot|$ is applied to achieve translation invariance.
In our work, the nonlinear layer aims to preserve local fractal dimension such that the local structural information of image will be enhanced.
The geometric invariance of feature is not our goal.
Finally, different from wavelet transformation, our fractal-based CNNs do not down-sample filtering result (i.e., pooling operation).

\section{Experiments}
\subsection{The AFM Image Benchmark and Protocols}
We apply our fractal dimension invariant filtering method to a challenging real-world task: detecting structural curves in AFM images of materials.
The demo code and partial data are in \url{https://sites.google.com/site/htxu313/resources/software}.
The images in this study are $40$ atomic force microscopy (AFM) phase images of nano-fibers. 
Each image is taken in tapping mode at a $10$ $\mu$m and with size $512\times 512$.
The fibrillar structure of the material has a huge influence on its electronic properties, which is represented via the complex salient curves in the image, as Fig.\ref{Fig1a} shows.
Detecting curves from the AFM images is challenging.
First, the AFM images often suffer from heavy noise and low contrast, which has negative influences on curve detection.
Second, the curves in these scenes are very complicated --- dense curves (i.e., nano fibers) with different shapes and directions are distributed in the image randomly and have overlaps with each other.
The ground truth of curves are extracted manually by a semi-automatic tool called FiberApp~\cite{usov2015fiberapp}.

We test our \textbf{FraCNN}-based curve detector with the original \textbf{FDIF}-based detector in both unsupervised and supervised cases.
Specifically, we consider these two detectors with thresholding-based binary processing (\textbf{BP}) and logistic regression (\textbf{LR}) as the last layer, respectively.
The size of filters used in FDIF and FraCNN is $9\times 9$, and the number of anisotropic filters used in FraCNN is $30$, as shown in Fig.~\ref{fig3}.
For investigating the influence of model's iteration number (depth) on learning results, we set the iteration number of FDIF to be $3$ (relatively shallow) or $6$ (relatively deep).
Accordingly, the depth of FraCNN is $6$ or $12$. 
In the supervised case (note only for last layer), we use $20$ AFM images as training set and the remaining $20$ AFM images as testing set. 
$80,000$ patches of size $9\times 9$ are sampled from the output images of FDIF or FraCNN to training parameters of the sigmoid layer. 
A half of training patches whose central pixels correspond to curves are labeled as positive samples, while the rest patches are negative ones.

For further demonstrating the superiority of our method, we consider the following competitors: the curve and line segment detector (\textbf{ELSD}) in~\cite{puatruaucean2012parameterless};
the traditional \textbf{Frangi} filtering-based curve detector~\cite{frangi1998multiscale}
the simple logistic regression \textbf{LR} using patches as features directly;
the classical CNN so-called \textbf{LeNet}~\cite{lecun1998gradient};
the state-of-art holistically-nested edge detector (\textbf{HED})~\cite{xie2015holistically}.
Although the HED is originally designed to detect edges, it should also be suitable to detect curves because both curves and edges satisfies the assumption of multi-scale consistency.
Therefore, we use the $20$ training images to fine-tune the pre-trained HED model and learn a curve detector accordingly.\footnote{The training code and pre-trained model is from \url{https://github.com/s9xie/hed}.}
Following the instruction in~\cite{xie2015holistically}, a post-process is applied to the output of CNNs, achieving the shrinkage and the binarization of detected curves.
The logistic regression is trained by $80,000$ patches with size $9\times 9$ sampled randomly from training images.
The training samples of the LeNet is also $80,000$ patches of images, the only difference is that the size of the patches is $28\times 28$.
In the testing phase, each patch of testing image will be classified and its label will be used as the corresponding pixel value of final binary map.

Similar to contour detection~\cite{arbelaez2011contour}, we use the standard metrics for curve detection, including the optimal F-score with fixed threshold (ODS), the optimal F-score with per-image best threshold (OIS), and average precision (AP).

\begin{table}[t!]
  \centering
  \small\caption{Performance comparison for various methods.\label{tab1}}
  \begin{threeparttable}[c]
    \begin{small}
      \begin{tabular}{
        @{\hspace{3pt}}c|l|c@{\hspace{3pt}}
        @{\hspace{3pt}}c@{\hspace{3pt}}
        @{\hspace{3pt}}c@{\hspace{3pt}}
        }
        \hline\hline
        \multicolumn{2}{c|}{Method} & ODS & OIS & AP\\ \hline
        \multirow{6}{*}{Non-Learning}&ELSD~\cite{puatruaucean2012parameterless}&0.058&0.058&0.030\\
        &Frangi~\cite{frangi1998multiscale} &0.629 &0.659 &0.578 \\
        &FDIF($\times 3$)+BP&0.717 &0.735 &0.699 \\
        &FDIF($\times 6$)+BP&0.715 &0.733 &0.695 \\
        &FraCNN($\times 6$)+BP&0.691 &0.719 &0.708 \\
        &FraCNN($\times 12$)+BP&0.689&0.715 &0.702 \\ \hline
		\multirow{7}{*}{Learning}&LR  &0.639  &0.706  &0.707  \\
		&LeNet~\cite{lecun1998gradient} &0.677 &0.718 &0.643 \\
		&HED~\cite{xie2015holistically} &0.722 &0.739 &\textbf{0.784} \\
		&FDIF($\times 3$)+LR &0.728 &0.770 &0.700 \\
		&FDIF($\times 6$)+LR &0.724 &0.767 &0.697 \\
		&FraCNN($\times 6$)+LR &\textbf{0.743} &\textbf{0.782} &0.730 \\
		&FraCNN($\times 12$)+LR &0.739 &0.774 &0.718\\
		\hline\hline
      \end{tabular}
    \end{small}
  \end{threeparttable}
\end{table}

\subsection{Experimental Results}
Table~\ref{tab1} gives comparison results for various methods, and Fig.~\ref{fig6} and \ref{fig8} visualize some typical results. More experimental results are attached in the appendix.

The traditional image processing methods like the ELSD and the Frangi filter seems unsuitable for detecting complicated curves in our case.
The ELSD method aims at detecting line segments and ellipse curves of rigid body in natural image.
The Frangi filter is originally designed for detecting vessels from medical images.
Both of these two methods can only detect sparse curves from relatively smooth background.
In our case, however, the curves of nano-fibers are very dense and complex and the AFM images are generally noisy.
As a result, the ELSD cannot detect complete curves and obtains very low ODS, OIS, and AP while the Frangi filtering method is not robust to noise and the change of contrast, which can only obtain chaotic results.


\begin{figure}[!t]
\begin{center}
\subfigure[AFM image]{
\includegraphics[width=0.22\linewidth]{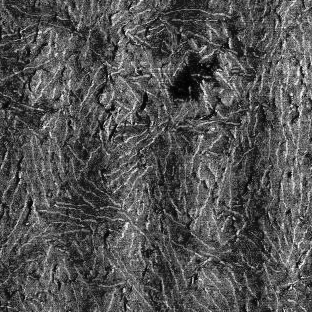}\label{Fig6a}
}
\subfigure[Manual labels]{
\includegraphics[width=0.22\linewidth]{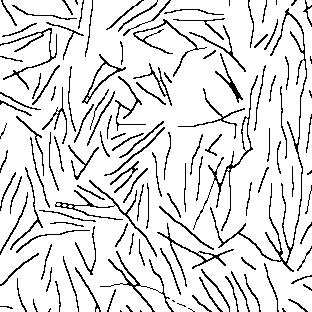}\label{Fig6b}
}
\subfigure[ELSD~\cite{puatruaucean2012parameterless}]{
\includegraphics[width=0.22\linewidth]{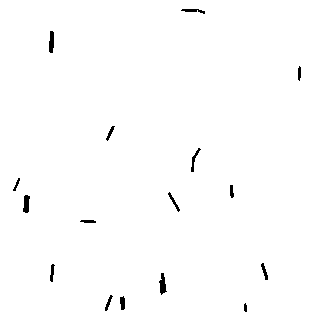}\label{Fig6c}
}
\subfigure[Frangi~\cite{frangi1998multiscale}]{
\includegraphics[width=0.22\linewidth]{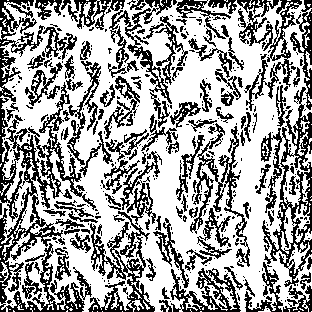}\label{Fig6d}
}
\subfigure[LR]{
\includegraphics[width=0.22\linewidth]{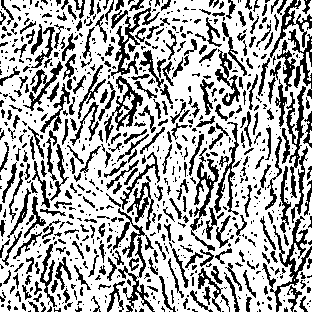}\label{Fig6e}
}
\subfigure[LeNet~\cite{lecun1998gradient}]{
\includegraphics[width=0.22\linewidth]{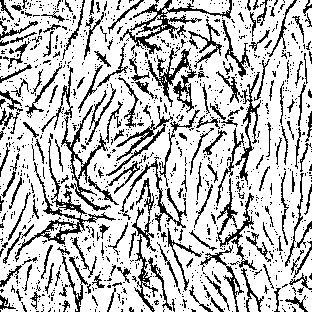}\label{Fig6f}
}
\subfigure[HED~\cite{xie2015holistically}]{
\includegraphics[width=0.22\linewidth]{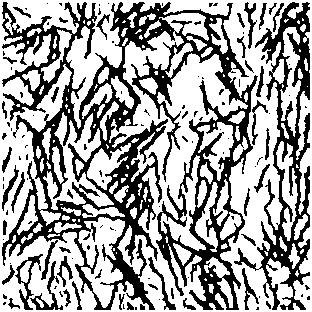}\label{Fig6g}
}
\subfigure[FDIF+BP]{
\includegraphics[width=0.22\linewidth]{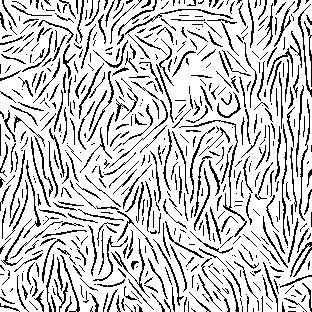}\label{Fig6h}
}\vspace{-10pt}
\end{center}
   \caption{Visual comparisons for various methods.}
\label{fig6}
\end{figure}


The learning-based approaches, including LR, LeNet, and HED, achieve much better results (i.e., higher ODS, OIS, and AP) than basic image processing methods.
However, their results are still very noisy. In Fig.~\ref{fig6}, LR's results contain many non-curve pixels and the many broken curves.
LeNet gets some improvements: long curves are detected correctly, but there are still many non-curve pixels.
HED is superior to LR and LeNet. Long curves are detected with more confidence and fewer incorrect isolated pixels appear in the results.
Table~\ref{tab1} shows the superiority of HED.

FDIF and FraCNN both achieve encouraging results. 
Specifically, our unsupervised methods, FDIF+BP and FraCNN+BP, outperform the other non-learning methods (ELSD and Frangi) notably, with better performance in Table~\ref{tab1} and visual results in Fig.~\ref{fig6}.
Additionally, FDIF+BP and FraCNN+BP are also better than some learning-based methods.
We can find that they get higher ODS, OIS and AP than LR and LeNet.
The comparison results still demonstrate that the fractal-based image model is suitable for the problem of curve detection, and our methods can extract representative features for curves.
In the supervised case, our FDIF+LR and FraCNN+LR methods outperform all the competitors in ODS and OIS while getting slightly worse AP than HED.
Moreover, from the enlarged comparison results in Fig.~\ref{fig8}, we can find that HED's result is still very coarse, while our method can get thin curves.
The results demonstrate that the proposed methods are at least comparable to the state-of-art in the problem of curve detection.
Note that our method is superior to HED in the aspect of computational complexity.
Specifically, in each layer, our FraCNN just applies $L$ 2D convolutions with kernel size $|B|$ to image $\bm{X}$, whose computational complexity is $\mathcal{O}(|\bm{X}||B|L)$, while HED applies $L$ 3D convolutions to an image tensor with $C$ channels, whose computational complexity is $\mathcal{O}(|\bm{X}||B|LC)$.

One important observation here is that although FraCNN can be viewed as an implementation of FDIF, it sometimes outperforms FDIF in Table~\ref{tab1}. A potential explanation for this phenomenon might be that FDIF is more sensitive to the noise in the image.
Specifically, the flexibility of FDIF on selecting directions might be a ``double-edged sword''.
Heavy noise in the image would lead to bad estimate of filter's direction and have negative influences on filtering results.
The FraCNN, however, uses a predefined anisotropic filter bank.
The limited options of directions might help to suppress the influence of noise.
Additionally, experimental results show that with the increase of iteration number and depth, the performance of our methods is degraded slightly.
In the viewpoint of numerical analysis, too many iterations or too deep architecture might lead to the underflow problem of pixel value. 
In the unsupervised case, instead of fine-tuning the threshold case by case, we uniformly set the threshold to $0.1$ for fair comparison. 
Note the threshold can have direct impact to final results: some underflow points might appear on curves, and thus the thresholding operation might break a complete curve into several pieces of short segments. 
In the supervised case, the underflow points in patches also hurt the representation of curve, which have negative influences on training the sigmoid layer.
\begin{figure}[!t]
\begin{center}
\subfigure[Manual labels]{
\includegraphics[width=0.3\linewidth]{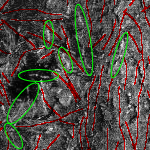}\label{Fig8a}
}
\subfigure[HED's curves]{
\includegraphics[width=0.3\linewidth]{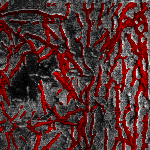}\label{Fig8b}
}
\subfigure[FDIF's curves]{
\includegraphics[width=0.3\linewidth]{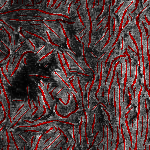}\label{Fig8c}
}\vspace{-10pt}
\end{center}
   \caption{Enlarged comparisons for various methods. The red curves are manually labeled results and the learning results of various methods. The green regions mark the unlabeled curves.}
\label{fig8}
\end{figure}

\begin{figure}[!t]
\begin{center}
\subfigure[]{
\includegraphics[width=0.14\linewidth]{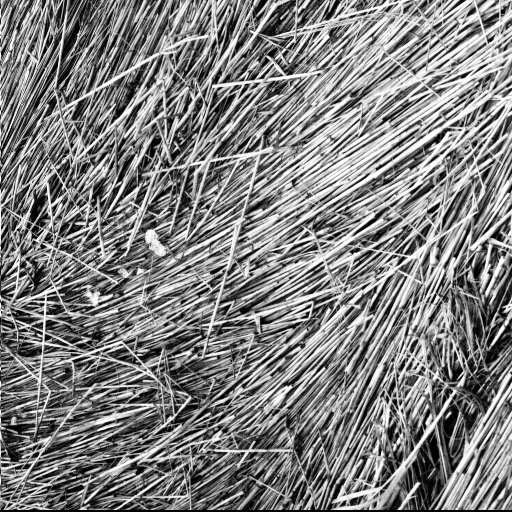}
\includegraphics[width=0.14\linewidth]{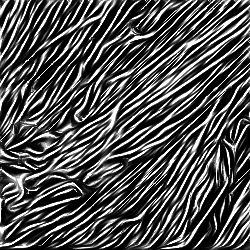}
}~
\subfigure[]{
\includegraphics[width=0.14\linewidth]{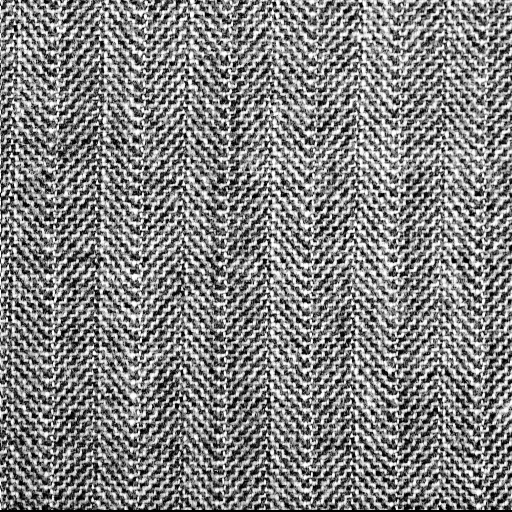}
\includegraphics[width=0.14\linewidth]{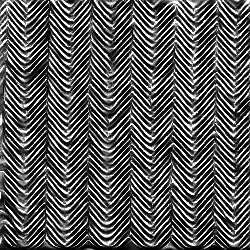}
}~
\subfigure[]{
\includegraphics[width=0.14\linewidth]{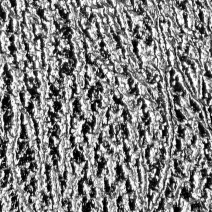}
\includegraphics[width=0.14\linewidth]{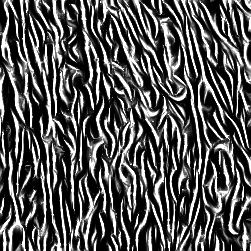}
}\vspace{-10pt}
\end{center}
\caption{Brodatz texture images and filtering results. 
The numerical results of FDIF($\times 3$)+LR are: OIS$=0.534$; ODS$=0.530$; AP$=0.803$. 
On the other hand, the results of HED (the best competitor) are: OIS$=0.522$; ODS$=0.518$; AP$=0.791$.
}
\label{figB}
\end{figure}

Furthermore, we select some texture images containing curves from the public Brodatz texture data set~\cite{brodatz1966textures}, label them manually, and test our method accordingly. 
Some typical visual results and numerical results are shown in Fig.~\ref{figB}, which further verify the performance of our method.

\subsection{Robustness to Missing Labels}
Compared with the state-of-art learning-based detector, an important advantage of the proposed method is that it is able to detect unlabeled curves.
The ground truth of curves is manually labeled. 
For labeling the texture-like complex image samples, humans are likely to miss some subtle or short curves in the labeling phase, as exemplified in Fig.~\ref{Fig8a}. As a result, the learning-based methods (e.g. HED) tend to ignore many existing curves or merge them together because in the training phase they have been ``taught'' to pay less attention to such unlabeled curves -- see Fig.~\ref{Fig8b}.
On the contrary, our method (e.g. FDIF+BP) is more robust to unlabeled curves -- see Fig.~\ref{Fig8c}. 
We think this is partially attributed to its intrinsic unsupervised learning nature: the representation of curve aims at preserving local fractal-dimension rather than approaching manual labels. 
As long as the response of a patch after anisotropic filtering is large enough, it will be preserved to represent curves.
In this viewpoint, our method can be utilized as a robust feature extraction method, which has potential to label salient curves automatically. 

\subsection{Other Possible Applications}
Besides curve detection, our fractal dimension invariant filtering method can also be used to create painting-style image from natural image. 
Considering the nature of most paintings that the objects in a painting are drawn via a series of curved strokes, we can treat paintings as a union of curves (fractals). 
Therefore, we can apply iterative FDIF method to natural image, enhancing their strokes and suppressing their textures.
Fig.~\ref{fig11} gives a typical example. 
More visual results are given in the appendix. 
Similar to the neural algorithm in~\cite{gatys2015neural}, our FDIF method has potential to generate diverse artistic styles via designing or learning different anisotropic filters. 
\begin{figure}[!t]
\begin{center}
\subfigure[Natural image]{
\includegraphics[width=0.4\linewidth]{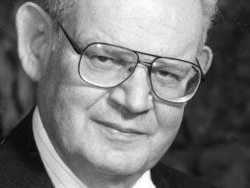}\label{Fig11a}
}
\subfigure[Painting-style image]{
\includegraphics[width=0.4\linewidth]{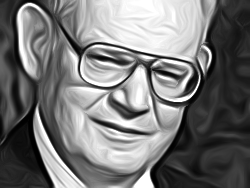}\label{Fig11b}
}\vspace{-10pt}
\end{center}
   \caption{The painting-style portrait of Benoit B. Mandelbrot, the author of ``\emph{The fractal geometry of nature}''~\cite{mandelbrot1983fractal}.}
\label{fig11}
\end{figure}

\section{Conclusion and Outlook}
Taking an image as a union of local fractals, this paper presents a model involving anisotropic filtering with fractal dimension preservation. 
The model is also re-implemented from a CNN interpretation. 
This work is the first attempt to bridge fractal-based image model with neural networks.

One notable character of our method is for its unsupervised feature extraction part, which does not rely on manually labeled data. 
This fact can be potentially of interest to the community: manual labeling in low-level vision problems is tedious and error-prone, which hurts the practical use of supervised learning approaches, while our method can obtain competitive performance on these task against supervised learning method (i.e., HED).
From the feature learning perspective, we believe that our fractal dimension invariant filtering can be further integrated with supervised learning techniques.
Additionally, we will further explore the potential applications of our method, e.g., the artistic style generation problem mentioned above. 

\textbf{Acknowledgment:} The work is supported in part via NSF IIS-1639792, NSF DMS-1317424, NSF DMS-1620345, NSF 1258425, NSFC 61471235, NSF FLAMEL IGERT Traineeship program, IGERT-CIF21, the Key Program of Shanghai Science and Technology Commission under Grant 15JC1401700, and the NSFC-Zhejiang Joint Fund for the Integration of Industrialization and Information under Grant U1609220. 

\section{Appendix}
\subsection{Proof of Theorem~\ref{the1}}
\begin{proof}
The mapping $g: \mathcal{A}\mapsto \mathcal{B}$ is bi-Lipschitz transform if and only if $g$ is invertible and there exists $0<c_1\leq c_2\leq \infty$ so that $c_1\|\bm{x}-\bm{y}\|\leq \| g(\bm{x})-g(\bm{y})\| \leq c_2\|\bm{x}-\bm{y}\|$ holds for all $\bm{x},\bm{y}\in \mathcal{A}$.
According to the definition, for arbitrary two points $\bm{x},\bm{y}\in \mathcal{F}$, we have
\begin{eqnarray*}\label{lip}
\begin{aligned}
&B_{\frac{r}{c_2}}(\bm{x})\subset B_{r}(g(\bm{x}))\subset B_{\frac{r}{c_1}}(\bm{x}),\\
&\mu(B_{\frac{r}{c_2}}(\bm{x}))\leq \mu(B_{r}(g(\bm{x})))\leq \mu(B_{\frac{r}{c_1}}(\bm{x})).
\end{aligned}
\end{eqnarray*}
Recall the relationship that $\log\mu(B_{r}(\bm{x}))=D\log{2r} +L$.
The following condition holds for all $r$'s:
\begin{eqnarray*}\label{bound}
\begin{aligned}
D\log\frac{2r}{c_2}+L\leq D_g\log{2r}+L_g \leq D\log\frac{2r}{c_1}+L.
\end{aligned}
\end{eqnarray*}
which implies $D_g=D$ and $L-D\log{c_2}\leq L_g\leq L-D\log{c_1}$.
\end{proof}

\subsection{Proof of Proposition~\ref{thm2}}
\begin{proof}
According to the assumption, the expectation of the filters in~(\ref{filter}) is $E_{\theta}(F_{\theta})=\int_0^{\pi}p_{\theta}F_{\theta}d\theta$, $p_{\theta}=\frac{1}{\pi}$. 
For each element $F_{\theta}(r, \phi)$, when $r=0$, we have
\begin{eqnarray*}
\begin{aligned}
E_{\theta}(F_{\theta}(0,\phi))=\int_0^{\pi}\frac{1}{\pi}F_{\theta}(0,\phi)d\theta
=\int_0^{\pi}\frac{1}{\pi |B|}d\theta=\frac{1}{|B|}.
\end{aligned}
\end{eqnarray*}
When $0<r\leq \sqrt{|B|}$, we have
\begin{eqnarray*}
\begin{aligned}
E_{\theta}(F_{\theta}(r,\phi))=\int_0^{\pi}\frac{1}{\pi}F_{\theta}(r,\phi)d\theta
=\lim_{\Delta\rightarrow 0}\frac{\int_{\phi-\Delta}^{\phi+\Delta}d\theta}{\pi |B|}=0.
\end{aligned}
\end{eqnarray*}
When $r>\sqrt{|B|}$, $E_{\theta}(F_{\theta}(r,\phi))=0$.
In summary, $E_{\theta}(F_{\theta})$ is the proposed impulse function.
\end{proof}

\subsection{More Enlarged Experimental Results}
Figs.~\ref{fig4}-\ref{fig10} show enlarged experimental results of detecting nano-fiber curves from AFM images. 
Figs.~\ref{fig11}-\ref{fig13} show painting-style generation results of several famous portraits. 
The contrast of each image in Figs.~\ref{fig11}-\ref{fig13} is adjusted, ensuring that the average intensity of FDIF's result is equal to the average intensity of original image. 

\begin{figure*}[!t]
\begin{center}
\subfigure[AFM image]{
\includegraphics[width=0.22\linewidth]{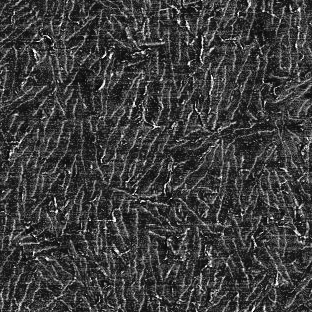}\label{Fig4a}
}
\subfigure[Manual labels]{
\includegraphics[width=0.22\linewidth]{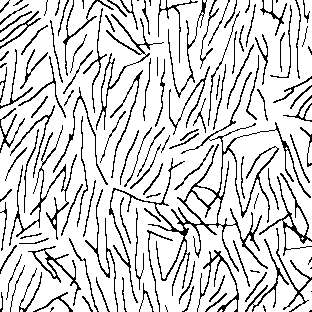}\label{Fig4b}
}
\subfigure[ELSD~\cite{puatruaucean2012parameterless}]{
\includegraphics[width=0.22\linewidth]{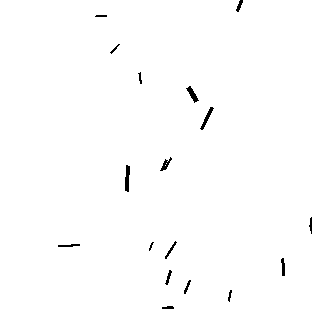}\label{Fig4c}
}
\subfigure[Frangi~\cite{frangi1998multiscale}]{
\includegraphics[width=0.22\linewidth]{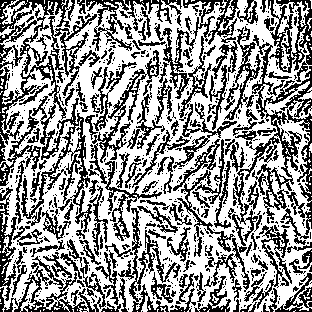}\label{Fig4d}
}
\subfigure[LR]{
\includegraphics[width=0.22\linewidth]{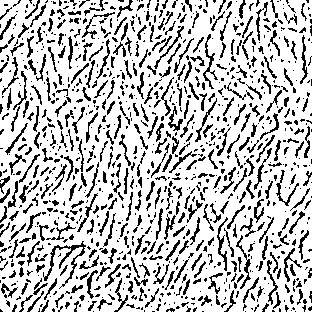}\label{Fig4e}
}
\subfigure[LeNet~\cite{lecun1998gradient}]{
\includegraphics[width=0.22\linewidth]{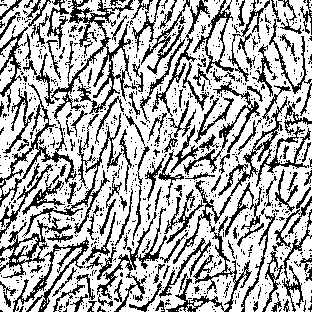}\label{Fig4f}
}
\subfigure[HED~\cite{xie2015holistically}]{
\includegraphics[width=0.22\linewidth]{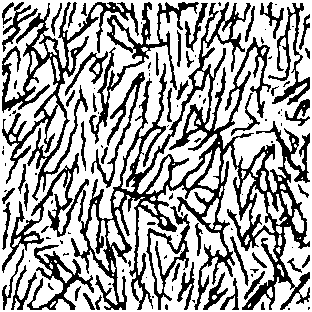}\label{Fig4g}
}
\subfigure[FDIF]{
\includegraphics[width=0.22\linewidth]{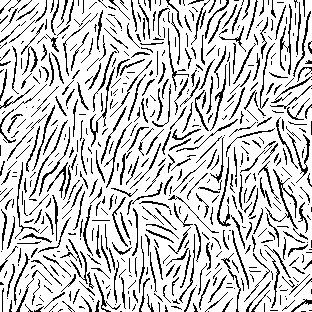}\label{Fig4h}
}
\end{center}
   \caption{Visual comparisons for various methods.}
\label{fig4}
\end{figure*}

\begin{figure*}[!t]
\begin{center}
\subfigure[AFM image]{
\includegraphics[width=0.22\linewidth]{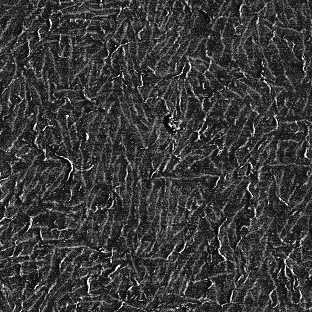}\label{Fig5a}
}
\subfigure[Manual labels]{
\includegraphics[width=0.22\linewidth]{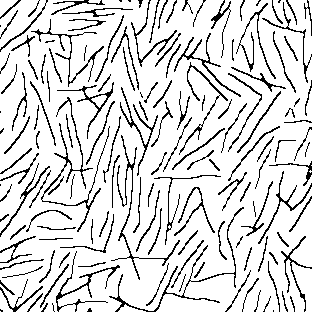}\label{Fig5b}
}
\subfigure[ELSD~\cite{puatruaucean2012parameterless}]{
\includegraphics[width=0.22\linewidth]{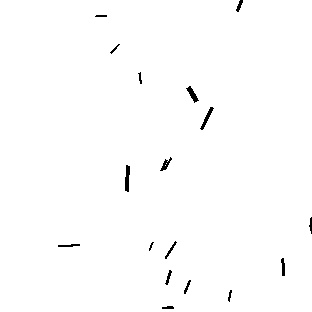}\label{Fig5c}
}
\subfigure[Frangi~\cite{frangi1998multiscale}]{
\includegraphics[width=0.22\linewidth]{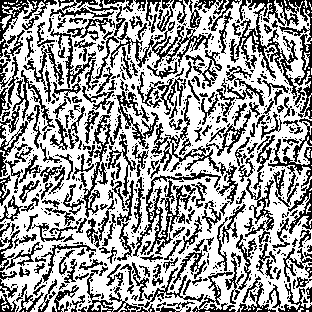}\label{Fig5d}
}
\subfigure[LR]{
\includegraphics[width=0.22\linewidth]{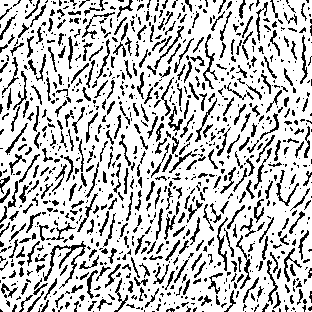}\label{Fig5e}
}
\subfigure[LeNet~\cite{lecun1998gradient}]{
\includegraphics[width=0.22\linewidth]{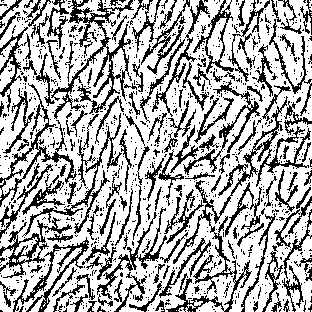}\label{Fig5f}
}
\subfigure[HED~\cite{xie2015holistically}]{
\includegraphics[width=0.22\linewidth]{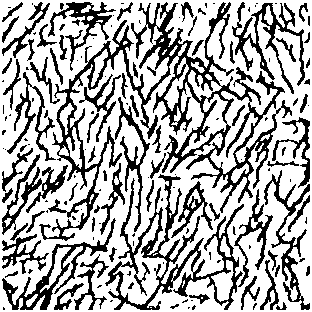}\label{Fig5g}
}
\subfigure[FDIF]{
\includegraphics[width=0.22\linewidth]{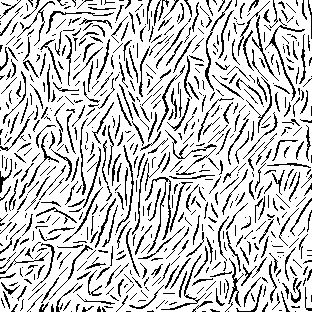}\label{Fig5h}
}
\end{center}
   \caption{Visual comparisons for various methods.}
\label{fig5}
\end{figure*}


\begin{figure*}[!t]
\begin{center}
\subfigure[AFM image]{
\includegraphics[width=0.22\linewidth]{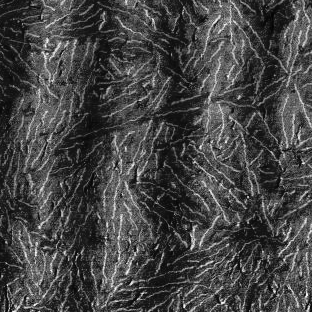}\label{Fig7a}
}
\subfigure[Manual labels]{
\includegraphics[width=0.22\linewidth]{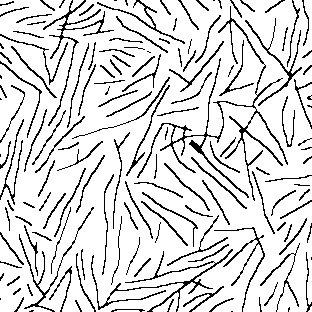}\label{Fig7b}
}
\subfigure[ELSD~\cite{puatruaucean2012parameterless}]{
\includegraphics[width=0.22\linewidth]{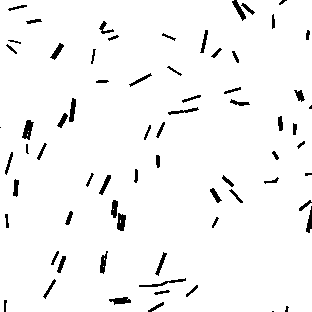}\label{Fig7c}
}
\subfigure[Frangi~\cite{frangi1998multiscale}]{
\includegraphics[width=0.22\linewidth]{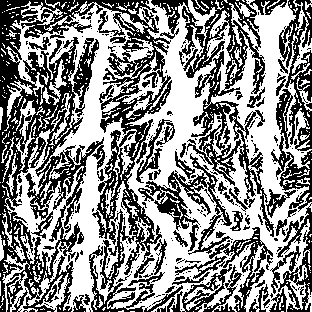}\label{Fig7d}
}
\subfigure[LR]{
\includegraphics[width=0.22\linewidth]{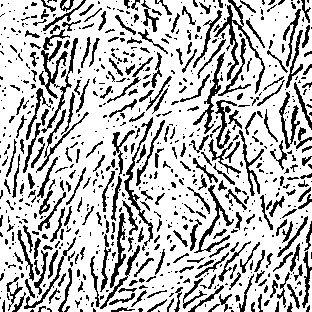}\label{Fig7e}
}
\subfigure[LeNet~\cite{lecun1998gradient}]{
\includegraphics[width=0.22\linewidth]{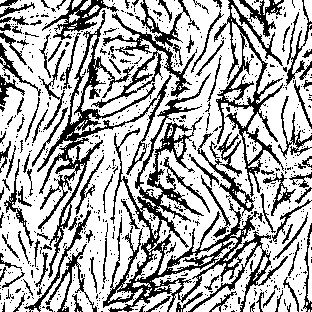}\label{Fig7f}
}
\subfigure[HED~\cite{xie2015holistically}]{
\includegraphics[width=0.22\linewidth]{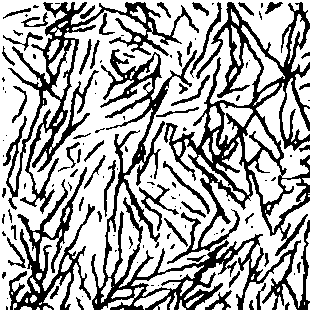}\label{Fig7g}
}
\subfigure[FDIF]{
\includegraphics[width=0.22\linewidth]{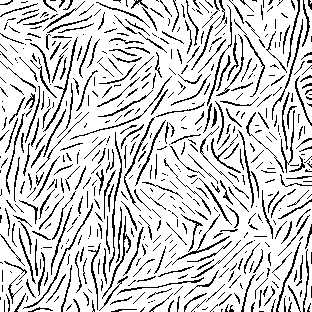}\label{Fig7h}
}
\end{center}
   \caption{Visual comparisons for various methods.}
\label{fig7}
\end{figure*}

\begin{figure*}[!t]
\begin{center}
\subfigure[AFM image]{
\includegraphics[width=0.22\linewidth]{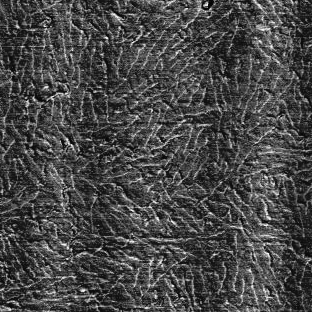}\label{Fig9a}
}
\subfigure[Manual labels]{
\includegraphics[width=0.22\linewidth]{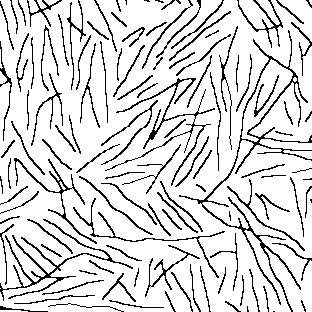}\label{Fig9b}
}
\subfigure[ELSD~\cite{puatruaucean2012parameterless}]{
\includegraphics[width=0.22\linewidth]{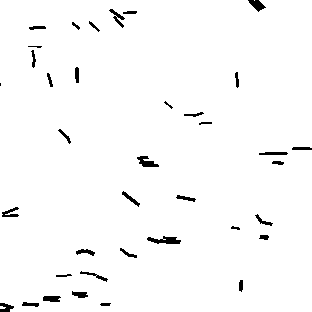}\label{Fig9c}
}
\subfigure[Frangi~\cite{frangi1998multiscale}]{
\includegraphics[width=0.22\linewidth]{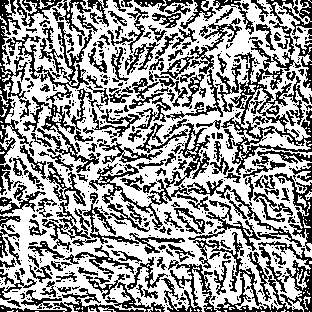}\label{Fig9d}
}
\subfigure[LR]{
\includegraphics[width=0.22\linewidth]{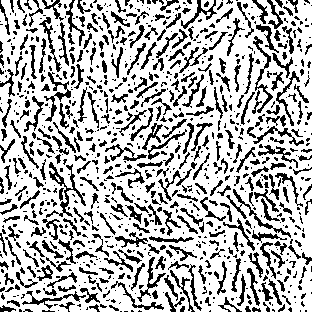}\label{Fig9e}
}
\subfigure[LeNet~\cite{lecun1998gradient}]{
\includegraphics[width=0.22\linewidth]{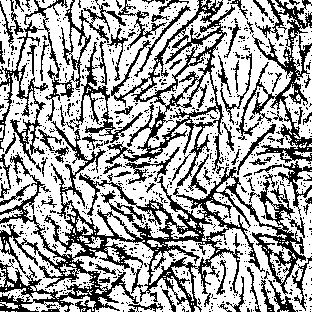}\label{Fig9f}
}
\subfigure[HED~\cite{xie2015holistically}]{
\includegraphics[width=0.22\linewidth]{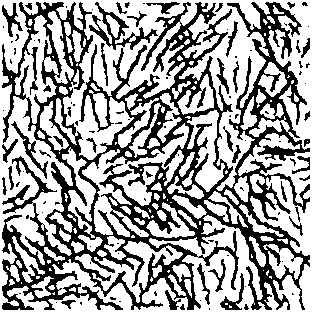}\label{Fig9g}
}
\subfigure[FDIF]{
\includegraphics[width=0.22\linewidth]{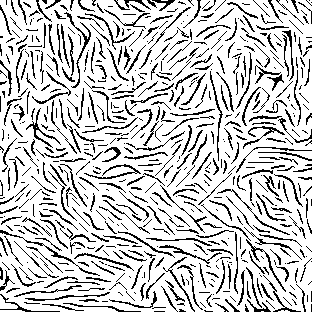}\label{Fig9h}
}
\end{center}
   \caption{Visual comparisons for various methods.}
\label{fig9}
\end{figure*}

\begin{figure*}[!t]
\begin{center}
\subfigure[AFM image]{
\includegraphics[width=0.22\linewidth]{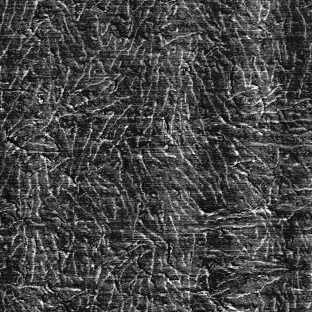}\label{Fig0a}
}
\subfigure[Manual labels]{
\includegraphics[width=0.22\linewidth]{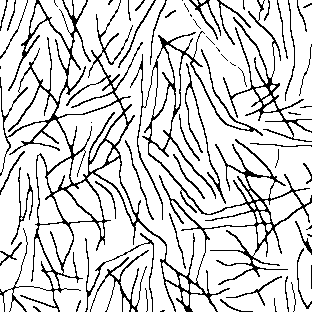}\label{Fig0b}
}
\subfigure[ELSD~\cite{puatruaucean2012parameterless}]{
\includegraphics[width=0.22\linewidth]{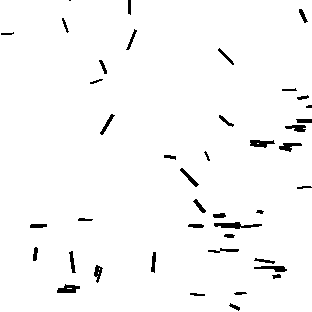}\label{Fig0c}
}
\subfigure[Frangi~\cite{frangi1998multiscale}]{
\includegraphics[width=0.22\linewidth]{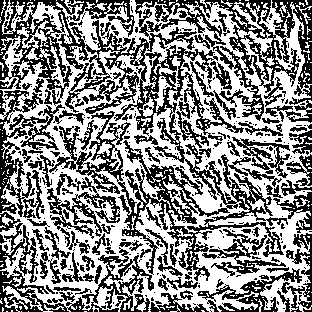}\label{Fig0d}
}
\subfigure[LR]{
\includegraphics[width=0.22\linewidth]{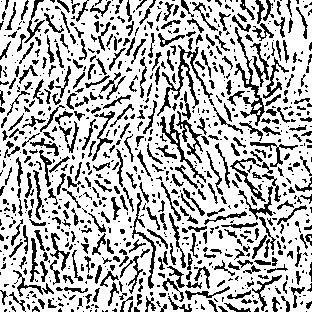}\label{Fig0e}
}
\subfigure[LeNet~\cite{lecun1998gradient}]{
\includegraphics[width=0.22\linewidth]{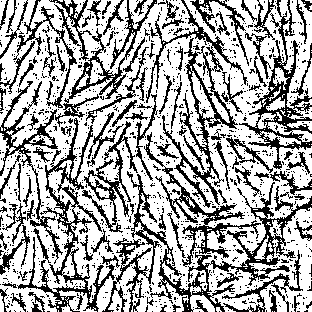}\label{Fig0f}
}
\subfigure[HED~\cite{xie2015holistically}]{
\includegraphics[width=0.22\linewidth]{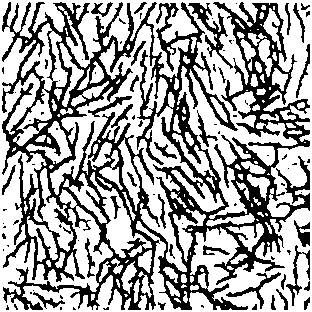}\label{Fig0g}
}
\subfigure[FDIF]{
\includegraphics[width=0.22\linewidth]{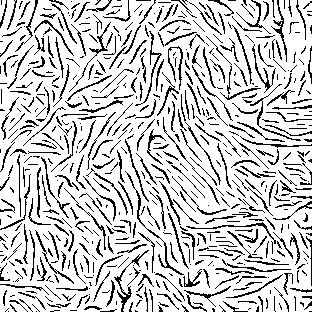}\label{Fig0h}
}
\end{center}
   \caption{Visual comparisons for various methods.}
\label{fig10}
\end{figure*}


\begin{figure*}[!t]
\begin{center}
\subfigure[Natural image]{
\includegraphics[width=0.22\linewidth]{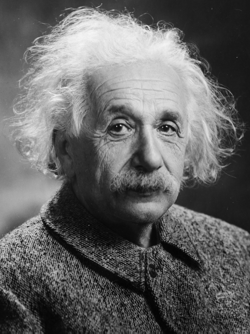}\label{Fig11a}
}
\subfigure[FDIF: \#1 Iteration]{
\includegraphics[width=0.22\linewidth]{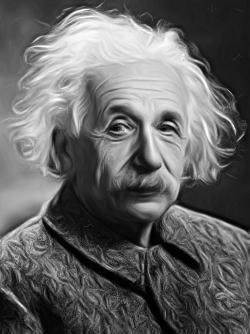}\label{Fig11b}
}
\subfigure[FDIF: \#2 Iteration]{
\includegraphics[width=0.22\linewidth]{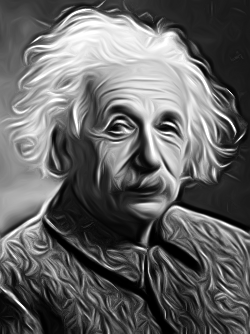}\label{Fig11c}
}
\subfigure[FDIF: \#3 Iteration]{
\includegraphics[width=0.22\linewidth]{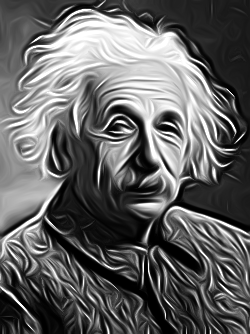}\label{Fig11d}
}\vspace{-10pt}
\end{center}
   \caption{The painting-style portrait of Albert Einstein generated via iterative FDIF method.}
\label{fig11}
\end{figure*}

\begin{figure*}[!t]
\begin{center}
\subfigure[Natural image]{
\includegraphics[width=0.22\linewidth]{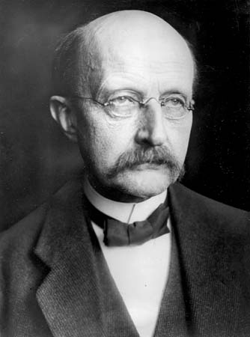}\label{Fig12a}
}
\subfigure[FDIF: \#1 Iteration]{
\includegraphics[width=0.22\linewidth]{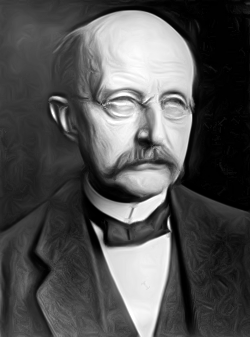}\label{Fig12b}
}
\subfigure[FDIF: \#2 Iteration]{
\includegraphics[width=0.22\linewidth]{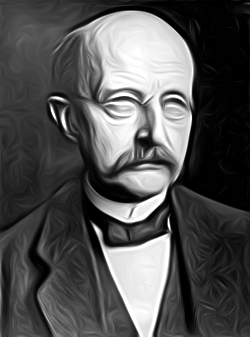}\label{Fig12c}
}
\subfigure[FDIF: \#3 Iteration]{
\includegraphics[width=0.22\linewidth]{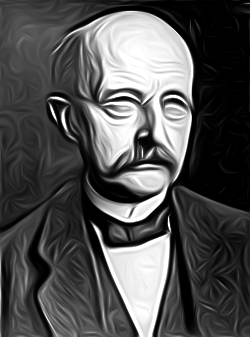}\label{Fig12d}
}\vspace{-10pt}
\end{center}
   \caption{The painting-style portrait of Max Planck generated via iterative FDIF method.}
\label{fig12}
\end{figure*}

\begin{figure*}[!t]
\begin{center}
\subfigure[Natural image]{
\includegraphics[width=0.22\linewidth]{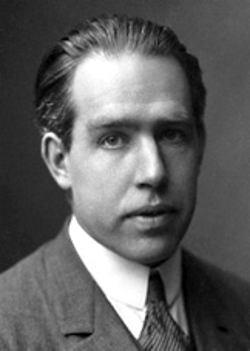}\label{Fig14a}
}
\subfigure[FDIF: \#1 Iteration]{
\includegraphics[width=0.22\linewidth]{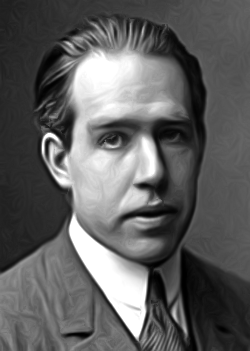}\label{Fig14b}
}
\subfigure[FDIF: \#2 Iteration]{
\includegraphics[width=0.22\linewidth]{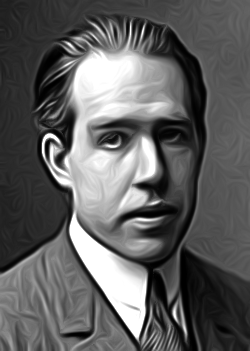}\label{Fig14c}
}
\subfigure[FDIF: \#3 Iteration]{
\includegraphics[width=0.22\linewidth]{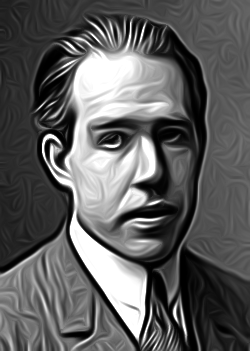}\label{Fig14d}
}\vspace{-10pt}
\end{center}
   \caption{The painting-style portrait of Niels Bohr generated via iterative FDIF method.}
\label{fig14}
\end{figure*}

\begin{figure*}[!t]
\begin{center}
\subfigure[Natural image]{
\includegraphics[width=0.22\linewidth]{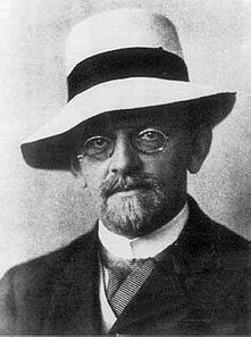}\label{Fig15a}
}
\subfigure[FDIF: \#1 Iteration]{
\includegraphics[width=0.22\linewidth]{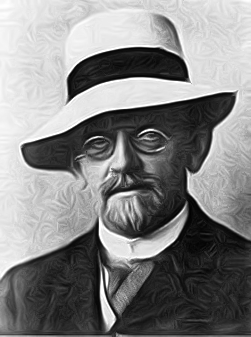}\label{Fig15b}
}
\subfigure[FDIF: \#2 Iteration]{
\includegraphics[width=0.22\linewidth]{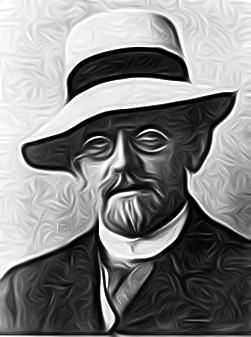}\label{Fig15c}
}
\subfigure[FDIF: \#3 Iteration]{
\includegraphics[width=0.22\linewidth]{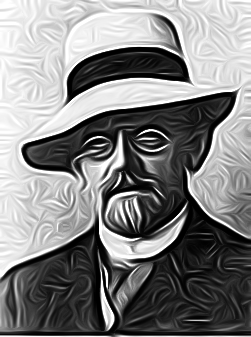}\label{Fig15d}
}\vspace{-10pt}
\end{center}
   \caption{The painting-style portrait of David Hilbert generated via iterative FDIF method.}
\label{fig15}
\end{figure*}

\begin{figure*}[!t]
\begin{center}
\subfigure[Natural image]{
\includegraphics[width=0.22\linewidth]{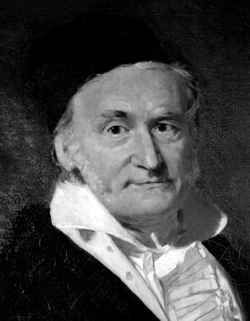}\label{Fig16a}
}
\subfigure[FDIF: \#1 Iteration]{
\includegraphics[width=0.22\linewidth]{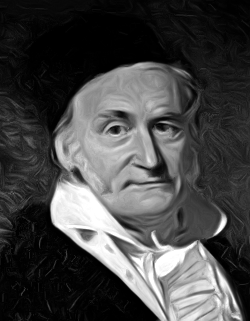}\label{Fig16b}
}
\subfigure[FDIF: \#2 Iteration]{
\includegraphics[width=0.22\linewidth]{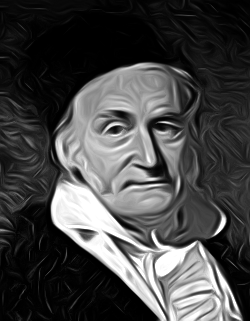}\label{Fig16c}
}
\subfigure[FDIF: \#3 Iteration]{
\includegraphics[width=0.22\linewidth]{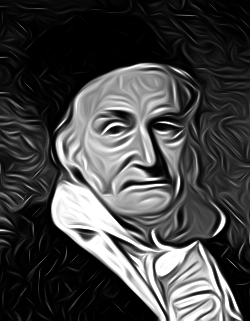}\label{Fig16d}
}\vspace{-10pt}
\end{center}
   \caption{The painting-style portrait of Carl Friedrich Gauss generated via iterative FDIF method.}
\label{fig16}
\end{figure*}

\begin{figure*}[!t]
\begin{center}
\subfigure[Natural image]{
\includegraphics[width=0.22\linewidth]{0_Mandelbrot.png}\label{Fig13a}
}
\subfigure[FDIF: \#1 Iteration]{
\includegraphics[width=0.22\linewidth]{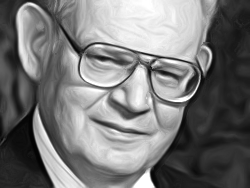}\label{Fig13b}
}
\subfigure[FDIF: \#2 Iteration]{
\includegraphics[width=0.22\linewidth]{2_Mandelbrot.png}\label{Fig13c}
}
\subfigure[FDIF: \#3 Iteration]{
\includegraphics[width=0.22\linewidth]{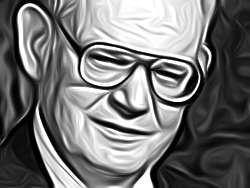}\label{Fig13d}
}\vspace{-10pt}
\end{center}
   \caption{The painting-style portrait of Benoit B. Mandelbrot generated via iterative FDIF method.}
\label{fig13}
\end{figure*}

{\small
\bibliographystyle{ieee}
\bibliography{egbib}
}

\end{document}